\documentclass[journal]{IEEEtran}
\usepackage{cite}

\usepackage{algorithm}
\usepackage{algorithmic}
\usepackage{graphicx}
\usepackage{textcomp}
\usepackage{epstopdf}
\usepackage{booktabs}
\usepackage{amsmath}
\usepackage{hyperref}
\usepackage{tipa}
\usepackage{esint}
\usepackage{arydshln}
\usepackage{multirow}
\usepackage{cite}
\usepackage{amssymb}
\usepackage{array}
\usepackage{color}
\usepackage{colortbl}
\usepackage[shortcuts]{extdash}
\usepackage{stfloats}
\usepackage{float}
\usepackage{makecell}

\usepackage[numbers,sort&compress]{natbib}
\usepackage[table,xcdraw]{xcolor}
\usepackage{newtxmath} %公式 字体
\usepackage{caption}
\usepackage{colortbl}
\usepackage{natbib}
\usepackage{amsmath}
\usepackage{bm}
\usepackage{amsfonts}
\usepackage{ulem}
\usepackage{soul}
\DeclareMathAlphabet\mathbfcal{OMS}{cmsy}{b}{n}
\usepackage{xspace}
\newcommand{\methodname}{LLM-DCP\xspace}
\def\wyf{\textcolor{black}}

\ifCLASSINFOpdf
\else
\fi

\makeatletter
\renewcommand{\maketag@@@}[1]{\hbox{\m@th\normalsize\normalfont#1}}%
\makeatother

\begin{document}

\title{Dynamic Compressing Prompts for Efficient Inference of Large Language Models}

\author{Jinwu Hu$^*$, Wei Zhang$^*$, Yufeng Wang, Yu Hu, Bin Xiao, \IEEEmembership{Senior Member, IEEE} \\ Mingkui Tan$^\dag$, \IEEEmembership{Senior Member, IEEE}, and Qing Du$^\dag$
\thanks{This work was partly supported by the National Natural Science Foundation of China under Grant 62072190.}
\thanks{Jinwu Hu and Wei Zhang are with the School of Software Engineering, South China University of Technology, and with Pazhou Lab, Guangzhou, China (e-mail: fhujinwu@gmail.com, zw2177738821@gmail.com).}
\thanks{Yufeng Wang is with the School of Future Technology, South China University of Technology, Guangzhou, China, and with Peng Cheng Laboratory, Shenzhen, China (e-mail: 202310193334@mail.scut.edu.cn).}
\thanks{Yu Hu is with the Department of Health Technology and Informatics, Hong Kong Polytechnic University, Hong Kong, China (e-mail: jasonscut@outlook.com).}
\thanks{Mingkui Tan and Qing Du are with the School of Software Engineering, South China University of Technology, Guangzhou, China (e-mail: mingkuitan@scut.edu.cn, duqing@scut.edu.cn).}
\thanks{Bin Xiao is with the Department of Computer Science and Technology, Chongqing University of Posts and Telecommunications, Chongqing, China (e-mail: xiaobin@cqupt.edu.cn).}
\thanks{$^*$Authors contributed equally. $^\dag$Corresponding authors}}

\maketitle

\begin{abstract}
Large Language Models (LLMs) have shown outstanding performance across a variety of tasks, partly due to advanced prompting techniques. However, these techniques often require lengthy prompts, which increase computational costs and can hinder performance because of the limited context windows of LLMs. While prompt compression is a straightforward solution, existing methods confront the challenges of retaining essential information, adapting to context changes, and remaining effective across different tasks. To tackle these issues, we propose a task-agnostic method called Dynamic Compressing Prompts (\methodname). Our method reduces the number of prompt tokens while aiming to preserve the performance as much as possible. We model prompt compression as a Markov Decision Process (MDP), enabling the DCP-Agent to sequentially remove redundant tokens by adapting to dynamic contexts and retaining crucial content. We develop a reward function for training the DCP-Agent that balances the compression rate, the quality of the LLM output, and the retention of key information. This allows for prompt token reduction without needing an external black-box LLM. Inspired by the progressive difficulty adjustment in curriculum learning, we introduce a Hierarchical Prompt Compression (HPC) training strategy that gradually increases the compression difficulty, enabling the DCP-Agent to learn an effective compression method that maintains information integrity. Experiments demonstrate that our method outperforms state-of-the-art techniques, especially at higher compression rates. The code for our approach will be available at https://github.com/Fhujinwu/DCP.
\end{abstract}


\IEEEpeerreviewmaketitle

\section{Introduction}
\begin{figure}[t] 
	\centering  
	\includegraphics[scale=0.30]{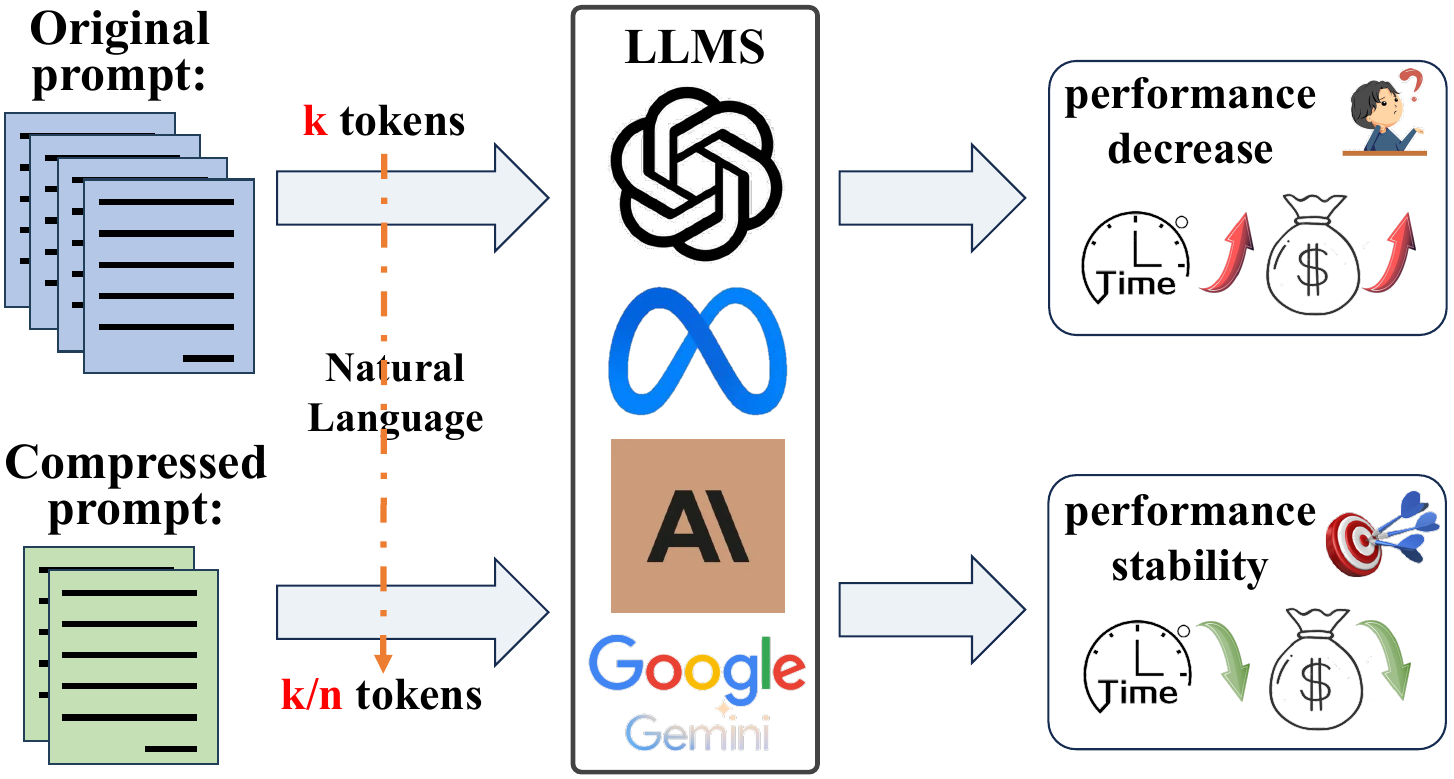}  
	\caption{Motivation for Prompt Compression of LLMs.}  
 \vspace{-6pt}
 \label{motivation}
\end{figure}

\IEEEPARstart{L}{arge} Language Models (LLMs) \cite{radford2018improving,brown2020language,zhang2022opt,achiam2023gpt,touvron2023llama-a,touvron2023llama-b} have shown excellent performance in different tasks, including recommender systems \cite{10506571} and drug design \cite{li2024empowering}. Many recently emerged prompting techniques for LLMs, such as Chain of Thought (CoT) \cite{wei2022chain}, Retrieval Augmented Generation (RAG) \cite{lewis2020retrieval}, Role-playing \cite{ouyang2022training}, etc., empower LLMs to handle complex and diverse tasks. However, these techniques increase the number of tokens required for the prompt, leading to additional computational and financial overhead, as well as reduced perceptual ability due to the limited context window of LLMs \cite{pan2024llmlingua} (see Fig. \ref{motivation}). While model quantization and expanding the context window can partially mitigate this issue, they do not fundamentally address the cost and performance limitations caused by long prompts. Consequently, prompt compression provides a straightforward solution aimed at shortening the original prompt while preserving key information and improving the LLM inference efficiency.

\textit{Unfortunately}, prompt compression presents several challenges, partly for the following reasons. \textit{1) Context sensitivity:} LLMs heavily rely on long prompts for context. Shortening prompts can negatively impact the ability of LLMs to generate coherent and accurate responses, requiring sophisticated compression techniques. \textit{2) Information retention:} Compressing prompts while preserving essential information is difficult. Key details can be lost during compression, leading to degraded performance in LLM outputs. \textit{3) Task-agnostic compression:} Developing a compression method adaptable across tasks, without being customized for specific scenarios, is particularly challenging due to the diverse nature of LLM applications.

To improve prompting efficiency, various prompt compression methods \cite{chevalier2023adapting,mu2024learning,xiao2023efficient,wingate2022prompt,jung2024discrete,jiang2023llmlingua,pan2024llmlingua,jiang2023longllmlingua,li2023compressing} have been explored, which can be broadly classified into white-box and black-box methods. The white-box compression method \cite{chevalier2023adapting,mu2024learning,xiao2023efficient,wingate2022prompt} compresses the prompt at the token-embedding level by modifying the model parameters, structure, and transformer self-attention mechanism. However, most high-performing LLMs (such as GPT-4 and Claude-3) are accessed through application programming interfaces (APIs), and the unavailability of source code severely limits the development and application of these methods. In response to the limited access to the source code of LLMs, black-box compression methods have emerged \cite{li2023compressing, jung2024discrete,jiang2023llmlingua,pan2024llmlingua,jiang2023longllmlingua}, leveraging the inherent redundancy of natural language \cite{shannon1951prediction}. The black-box compression method operates at the natural language level, aiming to shorten the original prompt without losing essential information. This method does not require access to the LLM source code for training or inference, reducing usage costs by directly minimizing input size. It also shortens inference time while preserving the performance of the LLM output.

Despite the recent black-box compression methods that can reduce the number of tokens in the prompt while maintaining the LLM output performance as much as possible, these methods still face certain limitations. \textbf{Firstly}, a common drawback of some existing task-aware compression methods \cite{jiang2023longllmlingua,jung2024discrete,xu2024recomp, shandilya2024taco} is that they are usually fine-tuned for specific tasks, and thus often difficult to use for different downstream task. For example, LongLLMLingua \cite{jiang2023longllmlingua} has to dynamically adjust the compression content according to the question, which may be difficult to use in summary tasks. \textbf{Secondly}, most task-agnostic methods \cite{li2023compressing, jiang2023llmlingua, pan2024llmlingua} estimate token importance using information entropy from causal language models, overlooking the sequential nature of prompt compression, where each token significance depends on the evolving context. \textbf{Thirdly}, many existing methods heavily depend on black-box LLMs during training, either for providing reward signals \cite{jung2024discrete, shandilya2024taco} or generating large-scale labeled data \cite{pan2024llmlingua}, leading to high training costs and limited practicality.

To address the above limitations, we propose a novel task-agnostic \textbf{D}ynamic \textbf{C}ompressing \textbf{P}rompts method, called \textbf{\methodname}, reducing the number of tokens of Prompt without affecting the output performance of LLMs as much as possible. Since the decision to remove or retain a token largely depends on the evolving context, we hypothesize that prompt compression can be viewed as a sequential decision-making process. In this process, redundancy is reduced iteratively while essential content is preserved, with each compression decision relying on the intermediate outcomes of previous iterations. Specifically, we model the prompt compression task as a Markov Decision Process (MDP), enabling the DCP-Agent to sequentially remove redundant tokens by adapting to dynamic contexts and retaining crucial content. Furthermore, we design a reward function for training the DCP-Agent that balances the compression rate, output distribution, and retention of key information, enabling prompt token reduction without compromising the LLM understanding and output. Importantly, this reward function does not require access to a black-box LLM, significantly reducing training costs. Additionally, inspired by curriculum learning \cite{wang2021survey,huang2022curriculum,10517422}, we introduce a Hierarchical Prompt Compression (HPC) training strategy that progressively increases the difficulty of compression, enabling the agent to effectively balance efficient compression with the protection of key information.

We summarize our main contributions as follows:
\begin{itemize}
    \item We propose a task-agnostic prompt compression method that models the compression process as a sequential decision-making problem using a Markov Decision Process (MDP). This method reduces the number of prompt tokens while aiming to minimize any negative impact on the LLM output performance. Experimental results show that LLM-DCP achieves approximately a 3.04\% improvement in Rouge-2 score over the state-of-the-art method, along with a higher compression ratio of 12.9x on the Arxiv-March23 dataset.
    \item To effectively train the DCP-Agent, we design a reward function that balances compression rate, output quality, and retention of key information. This reward function operates without direct supervision from the target LLM, significantly reducing training costs and enhancing practicality.
    \item We propose a Hierarchical Prompt Compression (HPC) training strategy that introduces progressively challenging compression tasks, allowing the proposed method to balance efficient compression with the preservation of key information effectively. Experiments show that the use of HPC yields a relative improvement of 25.5\% in compression ratio and 0.5 in $EM$ metric.
\end{itemize}

The remainder of this paper is organized as follows. Related work is presented in Section \ref{sec: related work}.  Section \ref{sec: definition} provides the problem definition and motivations. Section \ref{sec: method} describes the proposed \methodname. Section \ref{sec: experiment} provides the experiments and discussions. The conclusion of this paper is in Section \ref{sec: conclusion}.

\section{Related Work}
\label{sec: related work}
In this section, we focus on the content closely related to our work, which contains Large Language Models, Prompt Compression, and Reinforcement Learning.

\subsection{Large Language Models}

Large language models (LLMs) \cite{radford2018improving,brown2020language,zhang2022opt,achiam2023gpt,touvron2023llama-a,touvron2023llama-b,ouyang2022training}, such as the GPT series \cite{radford2018improving, brown2020language,ouyang2022training,achiam2023gpt}, have received significant attention for their excellent generalization and comprehension capabilities in natural language processing (NLP) such as multi-round dialogue \cite{achiam2023gpt, 10102573}, document summarisation \cite{9638337}, and question answering \cite{10637955}. A line of studies has attempted to enhance further the ability of LLMs to solve complex scenario tasks. Pan et al. \cite{10387715} and Yang et al. \cite{10417790} propose to use Knowledge Graph (KG) \cite{ji2021survey, 10234662} to enhance the reasoning power and interpretability of LLM. Wei et al. \cite{wei2022chain} propose the Chain of Thought (CoT) to strengthen the ability of LLMs to perform complex reasoning. Brown et al. \cite{brown2020language} propose In-Context Learning (ICL), where task-specific prompt templates are designed using a few labeled examples to guide the LLMs in generating predictions on new test data. Lewis et al. \cite{lewis2020retrieval} explore a Retrieval Augmented Generation (RAG) fine-tuning method that combines pre-trained parametric memory with non-parametric memory. However, many existing techniques for improving LLM capabilities have dramatically increased the length of the prompt. These rich and informative prompts can contain tens of thousands of tokens, which greatly increase inference time and application costs, and lead to poor performance due to the limited size of LLMs pre-training windows.

\subsection{Prompt Compression}

Prompts have become the dominant approach in NLP tasks \cite{brown2020language, schick2021exploiting, sanh2022multitask}, directly influencing the efficiency and performance of LLMs. As prompts grow longer, prompt compression has emerged as a crucial area of research \cite{lester2021power,liu2023pre}, with the goal of reducing LLM reasoning time and computational cost while maintaining performance. Prompt compression is typically categorized into two types: the white-box method and the black-box method \cite{wan2024efficient}.

The \textbf{white-box compression method} \cite{chevalier2023adapting,mu2024learning,xiao2023efficient,wingate2022prompt} reduces the prompt at the token-embedding level by modifying model parameters, architecture, and the self-attention mechanism of the Transformer.
Chevalier et al. \cite{chevalier2023adapting} propose AutoCompressors, which adapt pre-trained language models to compress long contexts into summary vectors that serve as soft prompts, improving language model performance and reducing inference costs. Mu et al. \cite{mu2024learning} train the gist model by modifying the attention mask to compress the prompt into a smaller set of ``gist" tokens to improve computational efficiency. Xiao et al. \cite{xiao2023efficient} propose StreamingLLM, a framework that enables large language models to handle infinite sequence lengths by utilizing attention sinks. Wingate et al. \cite{wingate2022prompt} proposed to learn a soft prompt compressed the context by aligning soft prompt-based model prediction with context-based predictive alignment. However, most high-performing LLMs (such as GPT-4 and Claude-3) are accessed via APIs, and the lack of access to source code significantly restricts the development and application of these methods. The \textbf{black-box compression method} \cite{li2023compressing, jung2024discrete,jiang2023llmlingua,pan2024llmlingua,jiang2023longllmlingua} operates at the natural language level, aiming to shorten the original prompt without losing essential information. Li et al. \cite{li2023compressing} propose Selective Context, which uses self-information to identify and prune redundant input, improving LLM reasoning efficiency by reducing memory costs and generation latency while maintaining task performance on long-context tasks.
Jiang et al. \cite{jiang2023llmlingua} propose LLMLingua, a coarse-to-fine prompt compression method that leverages a budget controller, a token-level iterative compression algorithm, and instruction tuning to achieve compression with minimal performance loss. Pan et al. \cite{pan2024llmlingua} create a task-agnostic data distillation procedure for better generalizability and efficiency. Jiang et al. \cite{jiang2023longllmlingua} propose LongLLMLingua to improve LLMs perception of key information for accelerated compression. Jung et al. \cite{jung2024discrete} employ a computationally efficient policy network that directly edits prompts. This type of method does not require access to the LLM source code for training or inference, reducing usage costs by directly minimizing input size.
  
\subsection{Reinforcement Learning} 

Reinforcement learning (RL) \cite{kaelbling1996reinforcement,9529088, 8844130, 9627535,10210487,9031418,hu2024dynamic} is a machine learning paradigm in which an agent interacts with its environment to achieve specific goals. In each interaction round, the agent takes an action based on the current state of the environment, receives feedback in the form of rewards or penalties, and subsequently updates its policy. The primary objective of RL is to maximize the cumulative reward. Unlike supervised learning, which aims to minimize the expected loss for a given data distribution, RL focuses on determining a strategy that maximizes the expected value of a reward function within a specified distribution. This trial-and-error approach to decision-making in uncertain environments allows RL to operate independently of labeled datasets.

To efficiently learn the optimal policy, RL has developed various algorithms. Sutton et al. proposed policy gradient \cite{sutton1999policy} algorithms, it puts the current state $s_t$ into the policy network $\bm{\pi}$ and outputs the current action $a_t$, the policy network $\bm{\pi}_\theta(a\mid s)$ is used directly to represent and control the behavior of an agent. Schulman et al. proposed proximal policy optimization \cite{schulman2017proximal} algorithms, where both the policy network $\bm{\pi}$ and the value evaluation model exist and introduce restrictions on the update magnitude, which is usually used for the training of LLMs. 
Recently RL has been developing rapidly in the field of LLM. Ouyang et al. \cite{ouyang2022training} point out that using reinforcement learning from human feedback (RLHF) \cite{christiano2017deep} can enable the model to follow a broad class of written instructions. Brohan et al. \cite{brohan2023can} introduce SayCan, which uses reinforcement learning as a method to learn language-conditioned value functions that provide guidance on what can happen in the real world, extracting and leveraging the knowledge of LLM in physical tasks to complete embodied tasks. Carta et al. \cite{carta2023grounding} propose GLAM, which uses the LLM as a policy that is incrementally updated as the agent interacts with the environment, and utilizes online reinforcement learning to improve the match between the LLM knowledge and the environment.

\begin{figure*}[ht] 
	\centering  
	\includegraphics[width=0.8\linewidth]{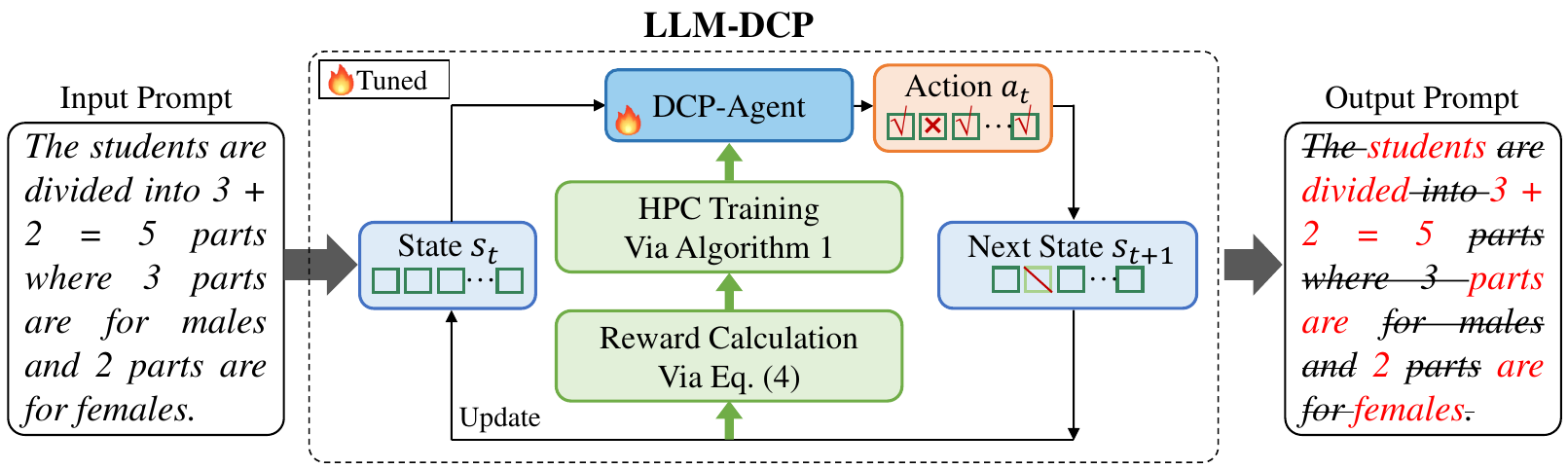}  
	\caption{ General diagram of proposed \methodname. \wyf{We model prompt compression as a Markov Decision Process (MDP) and train a DCP-Agent to determine an optimal compression pathway. The input prompt represented as a token sequence serves as the initial state of the MDP. At time step $t$, the DCP-Agent performs the action to select specific tokens to retain or discard, yielding a compressed token sequence as the next state $s_{t+1}$. Then the reward is calculated according to Eq. (\ref{equation: rt}). Our designed hierarchical prompt compression (HPC) training strategy collects the trajectory, which is applied to train the DCP-Agent. This process iterates until reaching the max trajectory length. The final token sequence is decoded into compressed text, with a much lower token number without affecting the output performance as much as possible.} }  
	\label{fig: method}
	\vspace{-6pt}    
\end{figure*} 

\section{Problem Definition and Motivations}
\label{sec: definition}
\subsection{Promblem Definition}

Given original prompt $\bm{x}=\{x_i\}_{i=1}^{L}$, a prompt compression system is designed to generate a compressed prompt $\widetilde{\bm{x}}=\{\widetilde{x}_i\}_{i=1}^{\widetilde{L}}$, where ${L}$ and $\widetilde{L}$ represent the numbers of tokens in $\bm{x}$ and $\widetilde{\bm{x}}$, respectively. The compression rate is defined as $\rho = \widetilde{L} / L$, $\rho \in [0, 1]$, and the compression ratio is $1/\rho$. \wyf{We prefer a smaller value of $\rho$ for lower inference cost.} Let $\bm{\widetilde{x}}_G$ represent the LLM-generated results derived by $\widetilde{\bm{x}}$ and $\bm{x}_G$ denotes the tokens derived by $\bm{x}$, the distribution of $\bm{\widetilde{x}}_G$ is expected to be as similar to $\bm{x}_G$ as possible. The objective of a prompt compression system can be formulated as:
\begin{equation}
\min_{\widetilde{\bm{x}}}
KL(P(\widetilde{\bm{x}}_G|\widetilde{\bm{x}}), P(\bm{x}_G|\bm{x}))+\rho,
\label{equ: problem definition}
\end{equation}
\subsection{Motivation}
Many existing prompt compression methods are task-aware, which limits their generalizability across different downstream tasks. Moreover,  most task-agnostic methods estimate token importance using information entropy from causal language models, neglecting the sequential nature of prompt compression, where each token significance depends on the evolving context. To address these issues, we hypothesize that prompt compression can be viewed as a dynamic, iterative decision-making process. Each compression step should reduce redundant information while leveraging the outcomes of previous steps to achieve efficient compression progressively. A natural idea emerges: \textit{Could we iteratively eliminate redundancy from the prompt while preserving its critical content through a series of decisions?}

The answer is yes. Inspired by trial-and-error learning, we model prompt compression as a Markov Decision Process (MDP), where the DCP-Agent iteratively compresses the prompt by removing redundant tokens while preserving essential content, with each decision building on the outcomes of previous steps for efficient, context-aware compression. We design a reward function that balances compression rate, output distribution, and key information retention, ensuring that the model understanding and output quality remain intact. Additionally, considering the challenges of retaining essential information while achieving high compression rates in the prompt compression task, we incorporate curriculum learning \cite{wang2021survey,huang2022curriculum,10174655}, progressively introducing more complex compression tasks to enhance the agent's ability to compress prompts efficiently while preserving essential content.

\section{Proposed Methods}
\label{sec: method}
In this paper, we propose a dynamic compressing prompts method, called \methodname, which seeks to remove redundant content in a given input prompt, thereby reducing computational cost and better using the limited context window in LLMs. As shown in Fig. \ref{fig: method}. \wyf{We model the prompt compression process as a Markov Decision Process (MDP) and train a DCP-Agent to determine an optimal compression pathway. Given an input prompt, we convert it to a token sequence, which serves as the initial state in the MDP framework. At time step $t$, the DCP-Agent selects specific tokens to be removed, yielding a compressed token sequence that constitutes the subsequent state $s_{t+1}$. Then the reward is calculated according to Eq. (\ref{equation: rt}). The trajectory is collected to train the DCP-Agent via our designed hierarchical prompt compression (HPC) training strategy. Additionally, the next state is input to the DCP-Agent for further iterations. This iterative process continues until the maximum trajectory length is reached. The final token sequence is decoded into compressed text, with a much lower token number without affecting the output performance.}

\subsection{Dynamic Compressing Prompts as an MDP}
We seek a general DCP-Agent to remove redundant tokens for a dynamic input prompt, thereby improving the inference efficiency while maintaining the quality of the generated text as much as possible. To this end, we formulate the step-by-step removal of redundant tokens as Markov Decision Process (MDP) \cite{van2012reinforcement}: $<\bm{S}, \bm{A}, \mathbfcal{T}, \mathbfcal{R}, \bm{\pi}>$. The state space of the environment is $\bm{S}$ and the action space of the agent is $\bm{A}$. At time step $t$, the agent takes the state $s_t\in \bm{S}$ as input and performs an action $a_t\in \bm{A}$ through the policy network $\bm{\pi}: \bm{S}\times \bm{A} \rightarrow \left[0,1\right]$. The environment changes to the next state $s_{t+1}=\mathbfcal{T}(s_{t}, a_{t})$ according to the transition function $\mathbfcal{T}$ and a reward $r_t=\mathbfcal{R}(s_{t},a_{t})$ is received with reward function $\mathbfcal{R}$. In this work, the MDP is detailed as follows:

\textbf{States} $\bm{S}$ \wyf{ is the description for the environment}. At time step $t$, the state is a compressed prompt: $s_t=\widetilde{x}_{t-1}=\{x_i\}_{i=1}^{\widetilde{L}_{t-1}}$, where $\widetilde{L}_{t-1}$ is the number of tokens after compression processing at time step $t-1$. Thus, the agent can predict which tokens need to be removed based on the current compressed prompt.

\textbf{Actions} $\bm{A}$ is a discrete set of actions the agent can take. In this task, the action space $\bm{A}=\{0,1\}^n$ is labeled for each token, with 0 indicating removal and 1 indicating preservation. At time step $t$, the agent gives the action $a_t \in \bm{A}$ based on the state $s_t$ to remove redundant tokens.

\textbf{Transition} $\mathbfcal{T}(\bm{S},\bm{A})$ is a function $\mathbfcal{T}$: $\bm{S} \times \bm{A} \rightarrow \bm{S}$ which maps a state $s_t$ into a new state $s_{t+1}$. When the maximum trajectory length is reached, this episode will terminated and $s_{T+1}$ is $None$. Otherwise, the action (preservation/removal) at time step $t$ for each token will result in a new prompt. It can be represented as:
\begin{equation}
    s_{t+1} = \mathbfcal{M}_{a_t}(s_t),
\end{equation}
where $\mathbfcal{M}_{a_t}(\cdot)$ is the operation that removes redundant tokens according to action $a_t$.

\textbf{Rewards} $\mathbfcal{R}(s_{t}, a_{t})$ is the reward function. In the LLM prompt compression task, the reward can be seen as minimizing the LLM output results while reducing the length of the prompt.
The details of the reward function we designed are given in the subsection \ref{subsec: reward}. 

\textbf{Policy} $\bm{\pi}_\theta(a\mid s): \bm{S}\times \bm{A} \rightarrow \left[0,1\right]$ describes the behaviors of the agent. During the training process, the agent takes the current state $s_t$ as input and outputs a probability distribution for each possible action $a_t \in \bm{A}=\{0,1\}^n$:
\begin{equation}
    \pi\left(a_t \mid s_t ; \theta\right)=\frac{\exp \left\{ f_{\theta}\left(s_t\right)_{i} \right\}}{\sum_{j=1}^N \exp \left\{ f_{\theta}\left(s_t\right)_{j} \right\}},
\end{equation}
where $f_{\theta}\left(s_t\right)$ is the output vector of the policy network with input $s_t$, and $i$ denotes the action style (0 or 1). The $\theta$ is the learnable parameter of the policy network.

\subsection{Reward function}
\label{subsec: reward}
Our goal is to reduce the number of tokens in the prompt without losing key information, not affecting LLM understanding of the prompt and the generation of results, as shown in Eq. (\ref{equ: problem definition}). Therefore, we design a reward function that takes into account the compression ratio, the Kullback-Leibler (KL) divergence \cite{kullback1951information} of the LLM-generated result distribution, and the degree of retention of key information from the prompt. The reward function is as follows:
\begin{align}
\label{equation: rt}
\mathbfcal{R}(s_{t}, a_{t}) &= \alpha \frac{1}{\rho} + \beta D(s_0, s_t) \nonumber \\
& - \gamma KL(P({s_t}_G|s_t), P({s_0}_G|s_0)) \nonumber \\
& - \mathbb{I}(\rho<c_s) P_{s} - \mathbb{I}(\rho>c_l)P_{l},
\end{align}
where $D(\cdot,\cdot)$ is used to compute the degree of key information retention for the original prompt (i.e., initial state $s_0$) and the compressed prompt (i.e., state $s_t$ at time step $t$) and here Bertscore \cite{zhang2020bertscore} is used, $c_s$ and $c_l$ are hyperparameters that indicate the lower and upper bounds of the expectation compression ratio, $P_{s}$ and $P_{l}$ are penalties for compressing prompts that are too short (over-compressed) and too long (under-compressed), respectively. The $\mathbb{I}(\cdot)$ is an indicator function. The ${s_0}_G$ and ${s_t}_G$ are the outputs of the LLM according to $s_0$ and $s_t$. Here the resulting distribution $P(\cdot)$ is not obtained from the target black-box LLM, but from a distribution-aligned small model, see subsection \ref{sub: Distribution Alignment} for details.

\textbf{Remark:} Unlike existing reinforcement learning-based summarization methods \cite{laban2020summary, ghalandari2022efficient}, the reward function we designed without considering the fluency and grammar of the compressed prompt, which is due to the fact that LLM has a good tolerance for prompts that lack fluency and grammatical errors \cite{jiang2023llmlingua, jiang2023longllmlingua, pan2024llmlingua}. Disregarding the fluency and grammar of the prompt is beneficial for obtaining a higher compression rate. In addition, the reward function we design does not require the involvement of a black-box LLM, which is different from the existing method \cite{jung2024discrete,shandilya2024taco}.

\subsection{Hierarchical Prompt Compression Training Strategy}
Considering the challenges of retaining essential information while achieving high compression ratio in the prompt compression task, and inspired by the progressive difficulty adjustment used in curriculum learning \cite{wang2021survey,huang2022curriculum}, we propose Hierarchical Prompt Compression (called \textbf{HPC}) training strategy for Proximal Policy Optimization (PPO) \cite{schulman2017proximal} process. The HPC training strategy introduces increasingly difficult compression tasks so that the agent gradually learns to balance efficient compression and preservation of key information. The details are as follows:

\textbf{Actor.} The actor (also called agent) $\pi_{\theta}$ is trained in binary classification (i.e., preservation or discarding of tokens) of the prompt according to the original prompt $\bm{x}=\{x_i\}_{i=1}^{L}$. To utilize the bidirectional contextual information of each token, we utilize the Transformer encoder as a feature extractor and then send the features to a linear classification layer. Specifically, at time step $t$, the state $s_t=\widetilde{x}_{t-1}=\{x_i\}_{i=1}^{\widetilde{L}_{t-1}}$ contains $\widetilde{L}_{t-1}$ tokens, which can be formalized as:
\begin{equation}
   \bm{h}=f_\theta(\widetilde{x}_{t-1}), 
\end{equation}
\begin{equation}
    p(x_i,\theta)=\text{softmax}(Wh_i+b),
\end{equation}
where $\bm{h}=\{h_i\}_{i=1}^{\widetilde{L}_{t-1}}$ is feature vectors for all tokens, $p(x_i,\theta) \in\mathbb{R}^2$ denotes the probability distribution of label $\{0,1\}$ for the $i-\text{th}$ token $x_i$. Here we use \textit{xlm-roberta-large} \cite{conneau-etal-2020-unsupervised} as Transformer encoder $f_\theta$. In the off-policy algorithm, the old policy $\pi_{\theta_{old}}$ with old parameters $\theta_{old}$ is used to collect trajectories with the environment, while the policy $\pi_{\theta}$ is updated using trajectories collected by $\pi_{\theta_{old}}$.

\begin{algorithm}[t]
    \small
    \caption{The HPC Training for \methodname}
    \label{alg:training_algorithm}
    \begin{algorithmic}[1]
    \REQUIRE
    The prompt for compression dataset $\mathcal{D}$, the DCP-Agent $\pi_{\boldsymbol{\theta}}$, the critic $V_\phi$ the reply buffer $\bm{\mathcal{B}}$, the maximum trajectory number of the buffer $M$, the iteration number of training $m$, the number of curriculum learning stages $P$ and the coefficients $c_s$ and $c_l$. 
    \STATE Initialize buffer $\bm{\mathcal{B}}$, actor parameters  $\theta$  and critic parameters $\phi$.
    \WHILE{Not convergence}
        \FOR{$P_i$ in $P$}
            \STATE Calculate $c_s$ and $c_l$ via Eq.(\ref{eq: HPC}).
            \FOR {$x_i$ in $\mathcal{D}$}
            
                \STATE Collect a trajectory $\tau=\{s_t,a_t,r_t,v_t,A^{\pi_{\theta_{old}}}(s_t, a_t)\}$ with old $\pi_{\theta_{old}}$ and $V_{\phi_{old}}$.
                
                \STATE Put $\tau$ into $\bm{\mathcal{B}}$.
                
                \IF {$length(\bm{\mathcal{B}}) == M$}
                    \FOR{$iteration=1, 2, \dots, M$}
                    
                        \STATE Uniformly  sample  $\tau \in \bm{\mathcal{B}}$.
                        
                        \STATE Calculate $\mathcal{J}(\theta)$ via Eq.(\ref{eq:Jtheta}).
    
                        \STATE Update $\theta$ to maximize $\mathcal{J}(\theta)$.
                        
                        \STATE Calculate TD error via Eq. (\ref{eq:TD}).
    
                         \STATE Update $\phi$ to minimize TD error $\delta_t$.
    
                    \ENDFOR
                    \STATE Empty the replay buffer $\bm{\mathcal{B}}$.
                    \STATE Update $\theta_{old} \leftarrow \theta$.
                    \STATE Update $\phi_{old} \leftarrow \phi$.
                \ENDIF
            \ENDFOR
        \ENDFOR
    \ENDWHILE
    \end{algorithmic}
\end{algorithm}

\textbf{Critic.} The critic $V_\phi(s)$ is used to estimate the expected return of the state $s_t$ and calculate the advantage, which can aid the actor in learning more efficiently and stably. Similar to the actor, the critic is composed of a pre-trained \textit{xlm-roberta-large} \cite{conneau-etal-2020-unsupervised} as an encoder, and with two Linear layers. Besides, the old critic $V_{\phi_{old}}(s)$ is used to collect trajectories, and the new critic $V_{\phi_{new}}(s)$ is updated using the collected trajectories.

\textbf{Learning Objectives.}
The goal of the learning is to maximize the expected long-term
return $\mathcal{J}(\theta)$:
\begin{align}
% \begin{split}
    \mathcal{J}(\theta)&=\mathbb{E}_{\tau \sim \pi_\theta(\tau)}[G(\tau)] \nonumber\\
    &=\mathbb{E}_{\tau \sim \pi_{\theta_{old}}(\tau)}    [\min (\delta A^{\pi_{\theta_{old}}}(s_t, a_t), \nonumber\\ &\operatorname{clip}\left(\delta, 1-\epsilon, 1+\epsilon\right) A^{\pi_{\theta_{old}}}(s_t, a_t))],
% \end{split}
\label{eq:Jtheta}
\end{align}
where $G(\tau)$ is the total return of the trajectory $\tau=\{s_t,a_t,r_t,v_t,A^{\pi_{\theta_{old}}}(s_t, a_t)\}$ obtained by $\pi_{\theta_{old}}$ and $V_{\phi_{old}}$, 
$\delta=\frac{\pi_\theta(a_t \mid s_t)}{\pi_{\theta_{old}}(a_t \mid s_t)}$ is the ratio of the probability of action $a_t$ given by  $\pi_\theta$ and $\pi_{\theta_{old}}$ for state $s_t$, and $\epsilon$ is a hyperparameter, we set to 0.15 in this paper. 
The operation $\operatorname{clip}\left(\delta, 1-\epsilon, 1+\epsilon\right)$ constrains $\delta$ to the range $\left[1-\epsilon, 1+\epsilon\right]$, and
$A^{\pi_{\theta_{old}}}(s_t, a_t)=r_t - V_{\phi_{old}}(s_t)$ is the advantage at $t$.

\textbf{HPC Training.}
The overview of the optimization process of the HPC training strategy is presented in Algorithm \ref{alg:training_algorithm}. Specifically, given a prompt for compression dataset $\mathcal{D}$, we use $\pi_{\theta_{old}}$ and $V_{\phi_{old}}(s)$ to interact with the environment to collect the trajectory $\tau=\{s_t,a_t,r_t,v_t\}$, and compute the advantage $A^{\pi_{\theta_{old}}}(s_t, a_t)$. 
During the collection of trajectories, the HPC training strategy increases the compression difficulty and guides the learning of the DCP-Agent incrementally by gradually decreasing the compression rate range $[c_s,c_l]$ (see Eq. \ref{equation: rt}) and the maximum trajectory length $T_{max}$ in stage ($P_i$). The $c_s$ and $c_l$ are adjusted as follows:
\begin{equation}
\left\{ \begin{array}{l}
c_s = 0.6 - (P_i + \frac{t}{T_{max}})\psi \\ 
c_l = 1.0 -  (P_i + \frac{t}{T_{max}})\psi
\end{array}, \right.
\label{eq: HPC}
\end{equation}
where $\psi$ set to 0.1, $P_i$ denotes the $i^{th}$ stage, with $i$ starting at 1 and $P_1$ = 1. Notably the learning stage size is set to 3, and $T_{max} = 2$ except for the third stage where $T_{max} = 1$.
\wyf{ This} easy to difficult curriculum learning strategy effectively improves the performance of prompt compression. 
We then put $\tau$ into the reply buffer $\bm{\mathcal{B}}$. 
When a certain number of trajectories (such as $M$) have been collected, they are used to train the actor and critic. In particular, we begin by uniformly sampling sequences from the replay buffer $\bm{\mathcal{B}}$, then compute the expected long-term return $\mathcal{J}(\theta)$ to optimize the policy parameters $\pi_{\theta}$. Additionally, the Temporal Difference (TD) error $\delta_t$ is calculated to refine the critic's parameters $V_{\phi}$:
\begin{equation}
    \delta_t=G_t-V_\phi\left(s_t\right),
    \label{eq:TD}
\end{equation}
where $G_t$ represents the total expected return starting from time step $t$. After conducting a certain number of training iterations using the samples from the existing replay buffer $\bm{\mathcal{B}}$, we clear the buffer and update the parameters of the old policy $\pi_{\theta_{old}}$ and critic $V_{\phi_{old}}$. This process is then repeated until convergence is achieved.

\begin{table*}[t]
\small
\centering
\renewcommand{\arraystretch}{1.1}
\renewcommand{\tabcolsep}{4.8pt}
\caption{Performance of different methods on the conversation (ShareGPT) and summarization (Arxiv-March23) tasks.}
    \begin{tabular}{lccccccccc}
    \hline \toprule[1.0pt]
    Method & Pub.'Year &BLEU $\uparrow$ &BLEURT $\uparrow$ &Rouge-1 $\uparrow$ 	&Rouge-2 $\uparrow$  &Rouge-L $\uparrow$ 	&BS F1 $\uparrow$ 	&Tokens  $\downarrow$ 	& $1/\rho$  $\uparrow$ \\ 
    \bottomrule[1.0pt] 
    \multicolumn{10}{c}{\cellcolor[HTML]{F2F2F2}\textbf{ShareGPT}}  \\ \hline
    Selective-Context \cite{li2023compressing} &EMNLP'2023	&38.53	&-0.21	&51.27	&38.35	&43.51	&78.30	&183	&3.3x \\
    LLMLingua\cite{jiang2023llmlingua}	&EMNLP'2023 &38.71	&-0.21	&51.43	&38.62	&43.57	&78.27	&186	&3.2x\\
    LLMLingua-2-small \cite{pan2024llmlingua}	&ACL'2024 &56.79	&0.37	&76.09	&58.47	&63.56	&89.54	&191	&3.1x\\
    LLMLingua-2 \cite{pan2024llmlingua}	&ACL'2024 &\underline{61.97} &\underline{0.47}	&\underline{78.64}	&\underline{63.07}	&\underline{67.50}	&\underline{90.87}	&\underline{184}	&\underline{3.3x}\\
    \textbf{\methodname (Ours)} &- &\textbf{64.93} &\textbf{0.54} &\textbf{80.24} &\textbf{65.54} &\textbf{69.89} &\textbf{91.80} &\textbf{175} &\textbf{3.4x} \\
    \bottomrule[1.0pt]
    \multicolumn{10}{c}{\cellcolor[HTML]{F2F2F2}\textbf{Arxiv-March23}}  \\ \hline
    Selective-Context \cite{li2023compressing} &EMNLP'2023	&8.83 &-0.61 &43.43 &13.46 &18.92 &73.75 &933 &11.8x \\
    LLMLingua\cite{jiang2023llmlingua}	&EMNLP'2023 &5.70 &-0.74 &32.29 &8.78 &15.17 &69.60 &1276 &8.7x\\
    LLMLingua-2-small \cite{pan2024llmlingua}	&ACL'2024 &8.56 &\textbf{-0.45} &45.52 &\underline{15.47} &\underline{21.09} &\underline{75.49} &1017 &10.9x\\
    LLMLingua-2 \cite{pan2024llmlingua}	&ACL'2024 &\textbf{10.84} &-0.57 &\underline{48.49} &14.62 &19.95 &75.15 &\underline{920} &\underline{12.0x}\\
    \textbf{\methodname (Ours)} &- &\underline{10.10} &\underline{-0.55} &\textbf{48.81} &\textbf{15.94} &\textbf{21.63} &\textbf{75.91} &\textbf{855} &\textbf{12.9x} \\
    \bottomrule[1.0pt]
    \end{tabular} 
\label{sharegpt_result}   
\end{table*}

\subsection{Distribution Alignment}
\label{sub: Distribution Alignment}
Due to the fact that the target black-box LLMs (e.g., GPT-4o-mini) are not available for the resulting distribution $P(\bm{\widetilde{x}}_G)$ generated by the compressed prompt $\widetilde{\bm{x}}$, we align a small model with the distribution of the target LLMs by instruction fine-tuning. Specifically, we use a pre-trained small language model $M_s$ for instruction tuning using the data pairs generated by black-box LLM of the target family. The optimization process of $M_s$ can be formulated as follows:
\begin{equation}
\min_{\theta_{\mathcal{M}_s}}\mathbb{E}\left[\frac{1}{N}\sum_{i=1}^N\mathcal{L}\left(\bm{x}_i,\bm{y}_{i,\text{LLM}};\theta_{\mathcal{M}_s}\right)\right],\end{equation}
where $\theta_{\mathcal{M}_s}$ is the parameters of $M_s$,  $(\bm{x}_i, \bm{y}_{i,\text{LLM}})$ is the pair of instruction $\bm{x}_i$ and the black-box LLM generated texts $\bm{y}_{i,\text{LLM}}$, and $N$ is the number of all examples used for instruction tuning. Notably, the Llama 3-8B \cite{dubey2024llama} is selected for $M_s$.

\section{Experiment}
\label{sec: experiment}
%总起句
In this section, we first introduce the experimental settings in subsection \ref{experimental settings}. Our proposed \methodname is compared against the state-of-the-art (SOTA) prompt compression methods in subsection \ref{comparison experiments}. We show relevant examples of the proposed \methodname in subsection \ref{examples}. We also performed numerous ablation experiments to validate the effectiveness of \methodname and to gain a deeper understanding of the proposed method in subsection \ref{ablation studies}. Additionally, we further discuss the effect of different hyperparameters on the proposed method in subsection \ref{discussion}.

\subsection{Experimental Settings}
\label{experimental settings}

\subsubsection{Compared methods}

Following the previous working setup \cite{pan2024llmlingua}, we compare the proposed \methodname with only three SOTA task-agnostic prompt compression methods.
\begin{itemize}
\item Selective-Context \cite{li2023compressing}: Use a small model to compute the self-information of each token and fuse it into a lexical unit \emph{u} (each lexical unit consists of multiple tokens $(x_t
, ..., x_{t+\alpha})$), retaining lexical unit self-information over a threshold value. 
\item LLMLingua \cite{jiang2023llmlingua}: It dynamically assigns different compression ratios ($\tau, \tau_{que}, \tau_{ins}, \tau_{dems}$) to the various parts of the prompt, divide the prompt into multiple segments $S=\{s_1, s_2, \ldots, s_m\}$, where tokens greater than threshold in each segment are retained.
\item LLMLingua-2 \cite{pan2024llmlingua}: It treats prompt compression as a token classification task, and it is available in LLMLingua-2-small and LLMLingua-2 versions.
\end{itemize}

\subsubsection{Datasets}

To comprehensively evaluate the effectiveness of the proposed \methodname, we conduct experiments on four different datasets on summarization, conversation, reasoning, and In-context learning (ICL) tasks.
\begin{itemize}
\item Arxiv-March23: It is a dataset consisting of the latest academic papers from the arXiv preprint repository, collected since March 2023. For our experimental evaluation, we employ a subset of 500 entries sourced from the dataset created by Li et al. \cite{li2023compressing}, which includes only the first two sections of each article to avoid excessive length.
\item ShareGPT: A dataset of 90k conversations collected from \textit{sharegpt.com}\footnote{\href{https://sharegpt.com/}{https://sharegpt.com/}}, involving multiple rounds of dialogue between users and ChatGPT in multiple languages and scenarios. We test the conversation task using sharegpt575 \cite{li2023compressing}, which contains 575 multi-round dialogue examples.
\item GSM8K \cite{cobbe2021gsm8k}: A widely used dataset for testing logic and mathematics in language modeling, containing 8.5k high-quality linguistically diverse mathematical problems. 
\item BBH \cite{suzgun2023challenging}: A subset of the BIG-Bench dataset \cite{srivastava2023beyond}, it focuses on a set of 23 challenging tasks covering multi-step arithmetic, algorithmic reasoning, language comprehension, and world knowledge. It is specifically designed to assess CoT prompting. For our experiments, we chose six tasks to test, including \textit{Boolean Expressions}, \textit{Causal Judgement}, \textit{Date Understanding}, \textit{Disambiguation QA}, \textit{Dyck Languages}, and \textit{Formal Fallacies}.
\end{itemize}

\subsubsection{Evaluation Metrics}
Following the settings of LLMLingua \cite{jiang2023llmlingua}, we use BLEU \cite{papineni2002bleu}, BLEURT \cite{sellam-etal-2020-bleurt}, ROUGE \cite{lin2004rouge} and BERTScore (BS F1) \cite{zhang2020bertscore} as evaluation metrics for Arxiv-March and ShareGPT. We use Exact Match ($EM$) \cite{jiang2023llmlingua} as a metric for GSM8K and BBH. In addition, the compression ratio ($1/\rho$) is also included in the assessment metrics to ensure fairness. Note that we use the tokenizer of Llama3\footnote{\href{https://huggingface.co/meta-llama/Meta-Llama-3-8B}{https://huggingface.co/meta-llama/Meta-Llama-3-8B}} to calculate the number of tokens.

\begin{table*}[t]
\small
\centering
\renewcommand{\arraystretch}{1.1}
\renewcommand{\tabcolsep}{13pt}
\caption{Performance of different methods on the reasoning (GSM8K), and In-context learning (BBH) tasks.}
    \begin{tabular}{lccccccc}
    \hline \toprule[1.0pt]
    \multirow{2}{*}{Method} & \multirow{2}{*}{Pub.'Year} & \multicolumn{3}{c}{\textit{1-shot constraint}} & \multicolumn{3}{c}{\textit{half-shot constraint}} \\
                &    & $EM\uparrow$      & Tokens  $\downarrow$      & $1/\rho$  $\uparrow$    & $EM\uparrow$       & Tokens  $\downarrow$       & $1/\rho$  $\uparrow$       \\
    \hline \toprule[1.0pt]
    \multicolumn{8}{c}{\cellcolor[HTML]{F2F2F2}\textbf{GSM8K}}  \\ \hline
    Selective-Context \cite{li2023compressing} &EMNLP'2023 &76.57 &436 &5.4x &76.15 &182 & 13.0x\\
    LLMLingua\cite{jiang2023llmlingua}	&EMNLP'2023 &76.72 &462 &5.1x &\underline{77.02} &174 &13.6x \\
    LLMLingua-2-small \cite{pan2024llmlingua}	&ACL'2024  &75.66 &425 &5.6x &76.80 &\underline{151} &\underline{15.7x} \\
    LLMLingua-2 \cite{pan2024llmlingua}	&ACL'2024 &\underline{76.87} &\underline{415} &\underline{5.7x} &76.80 &\textbf{140} &\textbf{16.9x} \\
    \textbf{\methodname (Ours)} &- &\textbf{77.03} &\textbf{343} &\textbf{6.9x} &\textbf{77.03} &153 &15.5x  \\
    \bottomrule[1.0pt]
    \multicolumn{8}{c}{\cellcolor[HTML]{F2F2F2}\textbf{BBH}}  \\ \hline
    Selective-Context \cite{li2023compressing} &EMNLP'2023 &\underline{82.81} &278 &2.8x &81.91 &152 &5.1x \\
    LLMLingua\cite{jiang2023llmlingua}	&EMNLP'2023 &81.68 &271 &2.9x &\textbf{84.72} &162 &4.8x \\
    LLMLingua-2-small \cite{pan2024llmlingua}	&ACL'2024  &82.73 &274 &2.8x &82.12 &155 &5.0x \\
    LLMLingua-2 \cite{pan2024llmlingua}	&ACL'2024 &82.41 &\underline{255} &\underline{3.0x} &82.64 &145 &5.3x \\
    \textbf{\methodname (Ours)} &- &\textbf{83.16} &\textbf{251} &\textbf{3.1x} &\underline{83.98} &\textbf{145} &\textbf{5.3x}  \\
    \bottomrule[1.0pt]
    \end{tabular} 
\label{gsm8k_bbh_result}   
\end{table*}
\subsubsection{Implementation Details}
Our proposed \methodname is implemented using PyTorch framework with Pytorch version 2.1.2 and runs on the 80G memory-sized NVIDIA A800 GPU with CUDA version 12.1. We use Adam as our optimizer to update the parameters of neural networks. The learning rate is set to $10^{-5}$ for the actor model and $10^{-6}$ for the critic model. The batch size is set to 4 and a total of 4 epochs are trained. The first and second stages are trained for 1 epoch respectively, and the third stage is trained for 2 epochs. The $P_s$ and $P_l$ in Eq. \ref{equation: rt} are set to 200 and 100, respectively.

For the training of model $M_s$ in subsection \ref{sub: Distribution Alignment}, we use the alpaca-gpt4-data\footnote{\href{https://huggingface.co/datasets/llm-wizard/alpaca-gpt4-data}{https://huggingface.co/datasets/llm-wizard/alpaca-gpt4-data}} dataset (randomly selected 80\% for the training set and 20\% for the test set) to fine-tune Llama3-8B, and the training framework used is LLaMA-Factory\footnote{\href{https://github.com/hiyouga/LLaMA-Factory}{https://github.com/hiyouga/LLaMA-Factory}}. Notably, the training hyperparameters use the default settings for full fine-tuning provided by LLaMA-Factory. We randomly selected 2048 prompt samples from the alpaca-gpt4-data dataset as training data to train the DCP-Agent. During the testing phase, we control the compression rate (e.g. 3x or 10x) by controlling the maximum step size. We employ the GPT-4o-mini-2024-07-18\footnote{\href{https://platform.openai.com/}{https://platform.openai.com/}} as the target LLMs, with greedy decoding at a temperature of 0 for enhanced stability across experiments.

\begin{figure*}[t] 
	\centering  
	\includegraphics[scale=0.36]{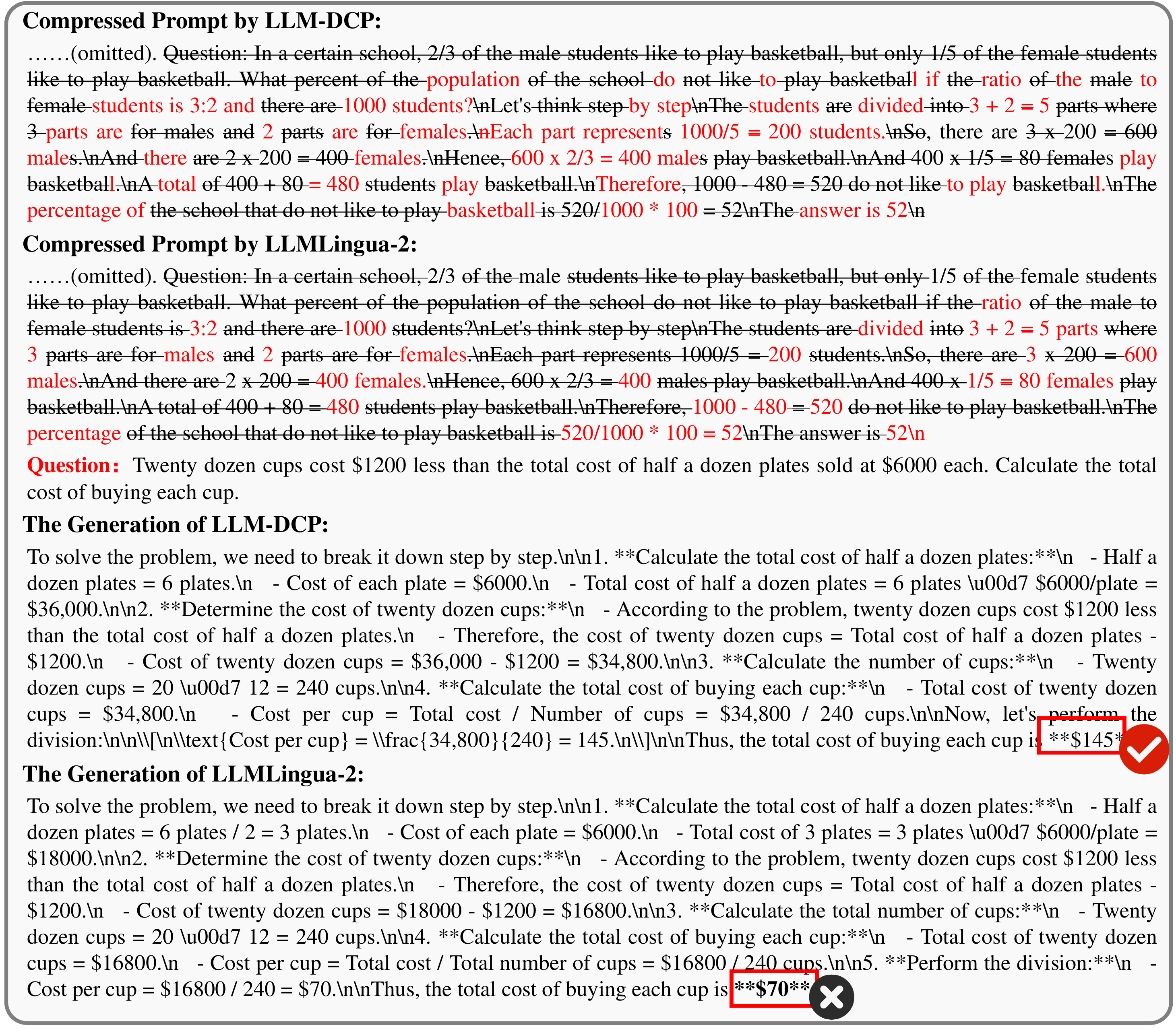}  
	\caption{Cases study on GSM8K dataset in \textit{1-shot constraint}. The \textcolor{red}{\textbf{red}} highlights the words that are preserved. The \sout{\textbf{strikethrough}} highlights the words that are removed.} 
	\label{fig: example}
	\vspace{-4pt}    
\end{figure*} 
\subsection{Comparison Experiments}
\label{comparison experiments}
We compare the proposed \methodname and three SOTA prompt compression methods to demonstrate the superior performance of our proposed method. We conduct experiments on a variety of downstream tasks, including conversation task (see Table \ref{sharegpt_result}), summarization task (see Table \ref{sharegpt_result}), reasoning task (see Table \ref{gsm8k_bbh_result}), and In-context learning task (see Table \ref{gsm8k_bbh_result}).

\textbf{Excellent performance of the \methodname in the conversation task.}
As shown in Table \ref{sharegpt_result}, \methodname outperforms other SOTA methods in the conversation task. Specifically, compared to LLMLingua-2, the proposed LLM-DCP improves about 4.8\% (61.97$\rightarrow$64.93) on BLEU and about 1.0\% (90.87$\rightarrow$91.80) on BS F1 at higher compression ratio (3.3x $\rightarrow$3.4x). The proposed \methodname achieves a 17.0\% relative improvement over the classical method, Selective-Context, on the BLEU metric. We conclude that \methodname removes redundant tokens according to prompt dynamic inputs, which allows outperforming SOTA methods in all metrics while maintaining a high compression ratio.

\textbf{Excellent performance of the \methodname in the summarization task.}
As shown in Table \ref{sharegpt_result}, our proposed \methodname outperforms SOTA methods in the summarization task. Specifically, the proposed \methodname achieves a relative improvement of 9.03\% (14.62$\rightarrow$15.94) on Rouge-2 metric compared to LLMLingua-2, while having a higher compression ratio (12.0x$\rightarrow$12.9x). Our proposed \methodname is not optimal in BLEU metric compared to LLMLingua-2, the main reason is that our DCP-Agent is trained on conversation data, while LLMLingua-2 is trained on the summarization task dataset, MeetingBank \cite{hu2023meetingbank}. Meanwhile, it exactly proves that the proposed \methodname still achieves better prompt compression performance in the cross-task situation.

\textbf{\methodname trade-off between performance and compression ratio.}
As shown in Table \ref{gsm8k_bbh_result}, the proposed \methodname outperforms the SOTA method in the reasoning task. Specifically, with the \textit{1-shot constraint}, our proposed \methodname has a relative improvement of 21.1\% (5.7x$\rightarrow$6.9x) in compression ratio and 0.2\% (76.87$\rightarrow$77.03) in $EM$ metric compared to LLMLingua-2. With \textit{half-shot constraint}, our proposed \methodname has a relative improvement of 0.3\% in $EM$ metric over LLMLingua-2, with a compression ratio of 15.5x. We conclude that the proposed \methodname trades off between performance and compression ratio. In the reasoning task, the performance of the $EM$ metric is not significantly improved between our proposed method and the existing prompt compression methods with approximately the same compression rate, a possible factor is that the target black-box model, GPT-4o-mini, already performs well on this task, even though the prompt of the CoT is not complete.

\textbf{Excellent performance of \methodname in In-context learning task.}
As shown in Table \ref{gsm8k_bbh_result}, the proposed \methodname outperforms the SOTA method in the $EM$ metric at a higher compression ratio. Specifically, with the \textit{1-shot constraint}, the proposed \methodname achieves a relative improvement of about 1.0\% (82.41$\rightarrow$83.16) in $EM$ metric compared to LLMLingua-2, along with a relative improvement of 3.3\% (3.0x$\rightarrow$3.1x) in compression ratio. With \textit{half-shot constraint}, the proposed \methodname improves the $EM$ metric by a relative 1.6\% (82.64$\rightarrow$83.98) compared to LLMLingua-2 while maintaining the same compression ratio. 

Overall, our proposed \methodname is a task-agnostic prompt compression method that achieves to outperform the SOTA method on four challenging tasks, such as the summarization task and reasoning task, by training only on the QA type dataset. On the one hand, it is because we model the prompt compression task as an MDP, and the DCP-Agent is able to remove redundant tokens according to the dynamic prompt inputs. On the other hand, it is because the reward function we designed balances the compression ratio, the output distribution of LLM, and the key information retention.

\subsection{Eaxmples of \methodname}
\label{examples}
We show an example of LLM-DCP and LLMLingua-2 on a reasoning task to demonstrate the effect of prompt compression, as shown in Fig.\ref{fig: example}. The \methodname and LLMLingua-2 are both tokens-level prompt compression methods, and although the compressed prompts are poorly readable, this does not have a significant impact on the understanding of the prompts by the LLM. In addition, our proposed \methodname retains more key information, which makes the prompts obtained after \methodname compression allow LLM to output more accurate answers compared to LLMLingua-2.

\begin{table}[t]
\small
\centering
\renewcommand{\arraystretch}{1.1}
\renewcommand{\tabcolsep}{9.2pt}
\caption{Ablation study on the GSM8K dataset with \textit{1-shot constraint}.}
    \begin{tabular}{lccc}
    \hline \toprule[1.0pt]
    \multirow{2}{*}{Version}  & \multicolumn{3}{c}{\textit{1-shot constraint}} \\
    & $EM\uparrow$      & Tokens  $\downarrow$      & $1/\rho$  $\uparrow$     \\
    \hline \toprule[1.0pt]
    Random	 &76.04 &428&5.5x  \\
    \methodname (w/o Training)	  &76.19 &\underline{422} &\underline{5.6x} \\
    \methodname (w/o HPC)	 &\underline{76.57} &431 &5.5x \\
    \textbf{\methodname (Ours)} &\textbf{77.03} &\textbf{343} &\textbf{6.9x}  \\
    \bottomrule[1.0pt]
    \end{tabular} 
\label{ablations_result}    
\end{table}

\subsection{Ablation Studies}
\label{ablation studies}
We follow the experimental setup of section \ref{experimental settings} and conduct a variety of ablation experiments to validate the effectiveness of modeling prompt compression as an MDP and the proposed HPC training strategy. Here, we experiment with the reasoning task in GSM8K dataset.

\textbf{Effectiveness of prompt compression with MDP.} We compare \methodname and random deletion tokens to demonstrate the effectiveness of modeling prompt compression as an MDP, as shown in Table \ref{ablations_result}. Compared to the random deletion tokens, the proposed LLM-DCP achieves a relative improvement of 1.3\% (76.04$\rightarrow$77.03) in $EM$ metric and a relative improvement of 25.5\% (5.5x$\rightarrow$6.9x) in compression ratio. A primary reason is the modeling of prompt compression as MDP, 
the trained DCP-Agent is able to iteratively refine the prompt by removing redundant tokens while preserving essential content, with each decision building on the outcomes of previous steps for efficient, context-aware compression.

\textbf{Effectiveness of HPC training strategy.} We compare \methodname and \methodname (w/o HPC) to verify the effectiveness of the proposed Hierarchical Prompt Compression training strategy, as shown in Table \ref{ablations_result}. Compared to \methodname (w/o HPC), \methodname has a relative improvement of 0.6\% (76.57$\rightarrow$77.03) in $EM$ metrics and a relative improvement of 25.5\% (5.5x$\rightarrow$6.9x) in the compression ratio. An important reason is that the HPC training strategy setting makes the training difficulty incremental step by step, which helps the DCP-Agent to better learn how to remove the redundant tokens in the dynamic prompt input.

\subsection{Discussion}
\label{discussion}
We conduct extensive experiments on the reasoning task to discuss further the effect of more details on the proposed method, such as the effect of each part of the reward function on the \methodname.

\begin{table}[t]
\small
\centering
\renewcommand{\arraystretch}{1.1}
\renewcommand{\tabcolsep}{12pt}
\caption{Experimental results for the component of the reward function on the GSM8K dataset with \textit{1-shot constraint}.}
    \begin{tabular}{cccccc}
    \hline \toprule[1.0pt]
    \multirow{2}{*}{$\alpha$} &\multirow{2}{*}{$\beta$} &\multirow{2}{*}{$\gamma$} & \multicolumn{3}{c}{\textit{1-shot constraint}} \\
    & & & $EM\uparrow$      & Tokens  $\downarrow$      & $1/\rho$  $\uparrow$     \\
    \hline \toprule[1.0pt]
    &$\checkmark$ &$\checkmark$	 &76.57 &\underline{339} &\underline{7.0x} \\
    $\checkmark$& &$\checkmark$	 &76.70 &396 &6.0x \\
    $\checkmark$ &$\checkmark$ &  &\underline{76.72}	  &\textbf{323} &\textbf{7.3x} \\
    $\checkmark$ &$\checkmark$ &$\checkmark$	 &\textbf{77.03} &343 &6.9x \\
    \bottomrule[1.0pt]
    \end{tabular} 
    \label{reward_result}   
\end{table}

\begin{figure}[t] 
	\centering  
	\includegraphics[scale=0.52]{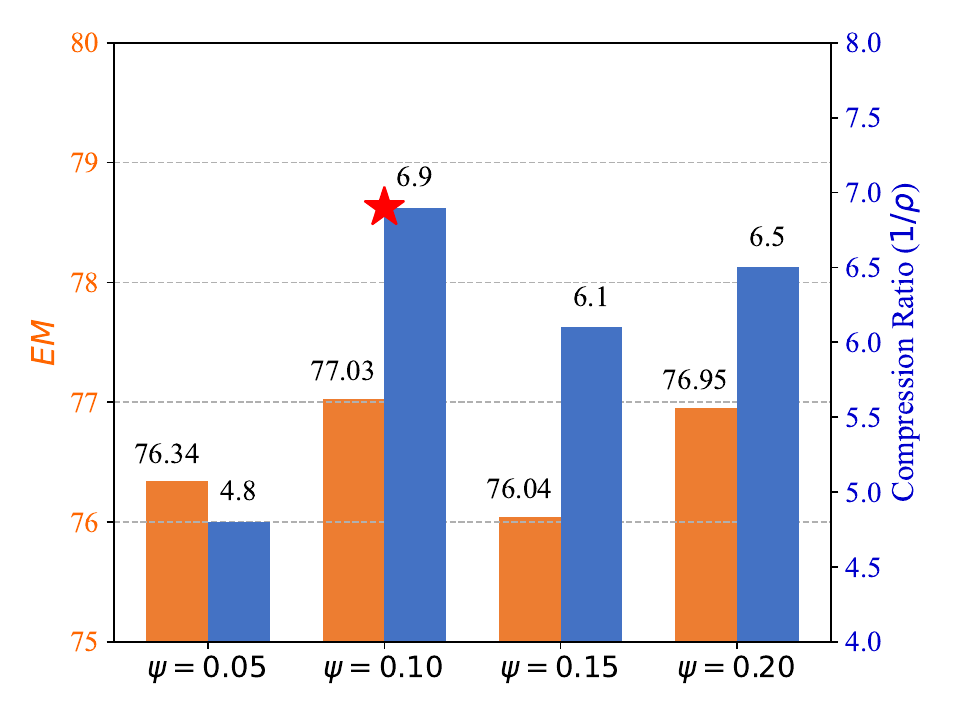}  
	\caption{Experimental results for different values of $\psi$ on the GSM8K dataset with \textit{1-shot constraint}.}  
	\label{fai_result}
\end{figure} 

\textbf{Effect of the reward function $\mathbfcal{R}$.}
We study the effects of compression ratio, LLM output distribution, and information retention in our proposed reward function on the performance of the proposed \methodname by adjusting $\alpha=0$, $\beta=0$ or $\gamma=0$ in Eq. \ref{equation: rt}. As shown in Table \ref{reward_result}, when $\alpha=0$, the lack of compression ratio of the reward signal may cause the DCP-Agent to delete some key information, leading to a decrease in the $EM$ metric. When $\beta=0$, the reward signal lacks the reward of key information retention, which may cause the DCP-Agent not to pay attention to the key information retention of the prompt before and after compression. In addition, when $\gamma = 0$, the reward signal lacks attention to the effect of the prompt on the LLM output distribution before and after compression, leading to a decrease in the $EM$ metric. In summary, the reward function we designed takes into account the compression ratio, the KL distribution of the LLM output, and the retention of key information, so that the trained DCP-Agent balances the compression ratio and the performance.

\textbf{Effect of the $\psi$ in Eq. \ref{eq: HPC}.} To study the effect of $\psi$ in Eq. \ref{eq: HPC} on the performance of \methodname in the proposed HPC training strategy, we take different values of $\psi$ and conduct experiments in the reasoning task. As shown in Fig. \ref{fai_result}, the performance of \methodname is optimal when $\psi=0.10$. With $\psi = 0.05$, the range of compression ratios is smaller, resulting in a compression ratio of only 4.8x. When $\psi \geq 0.10$, the compression ratios are all in a more appropriate range, but when $\psi=0.15$ and $\psi=0.20$, the compression ratios vary slightly more during the training process of HPC, which leads to the difficulty of learning the best compression strategy for the DCP-Agent.

\textbf{Performance on different target LLMs.}
We study the performance of the proposed \methodname and SOTA methods on GLM-4-Plus\footnote{\href{https://bigmodel.cn/}{https://bigmodel.cn}} with the GSM8K dataset. As shown in Table \ref{glm4_result}, the comparison results of our proposed \methodname and SOTA methods on $EM$ metric under \textit{half-shot constraint} present consistency with the target LLM as GPT-4o-min. Therefore,  \methodname is applicable to different black-box LLMs.

\begin{table}[t]
\small
\centering
\renewcommand{\arraystretch}{1.1}
\renewcommand{\tabcolsep}{2.6pt}
\caption{Experimental results for GLM-4-Plus on the GSM8K dataset with \textit{half-shot constraint}.}
    \begin{tabular}{lcccc}
    \hline \toprule[1.0pt]
    \multirow{2}{*}{Method}  & \multirow{2}{*}{Pub.'Year} & \multicolumn{3}{c}{\textit{half-shot constraint}} \\
    & &$EM\uparrow$      & Tokens  $\downarrow$      & $1/\rho$  $\uparrow$     \\
    \hline \toprule[1.0pt]
    Selective-Context \cite{li2023compressing} &EMNLP'2023 &77.79 &182 &13.0x  \\
    LLMLingua\cite{jiang2023llmlingua}	&EMNLP'2023 &\underline{79.07} &174 &13.6x \\
    LLMLingua-2-small \cite{pan2024llmlingua}	&ACL'2024  &77.63 &\underline{151} &\underline{15.7x}\\
     LLMLingua-2 \cite{pan2024llmlingua}	&ACL'2024 &76.19 &\textbf{140} &\textbf{16.9x}\\
    \textbf{\methodname (Ours)} &- &\textbf{79.76} &153 &15.5x  \\
    \bottomrule[1.0pt]
    \end{tabular} 
\label{glm4_result}  
\end{table}

\section{Conclusion}
\label{sec: conclusion}
In this paper, we present \methodname, a novel task-agnostic approach for prompt compression in Large Language Models (LLMs), aimed at reducing the number of tokens while maintaining output quality. We model the prompt compression task as a Markov Decision Process (MDP), enabling the DCP-Agent to iteratively compress the prompt by removing redundant tokens while preserving essential content, with each decision building on the outcomes of previous steps for efficient, context-aware compression. A carefully design reward function is introduced to balance compression rate, output distribution, and key information retention, ensuring effective compression without compromising LLM performance. Furthermore, we propose the Hierarchical Prompt Compression (HPC) training strategy, which employs a progressive training scheme to gradually increase compression difficulty, allowing the agent to learn an efficient compression strategy. We conduct experiments on a variety of downstream tasks, including the conversation task, the summarization task, the reasoning task, and the In-context learning task. Experiments demonstrate that our method performs better at higher compression rates than state-of-the-art methods.

\footnotesize
\bibliographystyle{IEEEtran}
\bibliography{reference}

\end{document}